\documentclass{article}

   \usepackage[preprint]{neurips_2024}

\usepackage{natbib}
\setcitestyle{numbers,sort&compress}
\usepackage[utf8]{inputenc} 
\usepackage[T1]{fontenc}    
\usepackage{hyperref}       
\usepackage{url}            
\usepackage{booktabs}       
\usepackage{amsfonts}       
\usepackage{nicefrac}       
\usepackage{microtype}      
\usepackage{xcolor}         
\usepackage{float}
\usepackage{mathtools}
\usepackage{times}
\usepackage{graphicx}
\usepackage{amssymb}
\usepackage{tabularx}
\usepackage{multirow}
\usepackage{algorithm} 
\usepackage{algpseudocode}
\usepackage{enumitem}
\setlist{nolistsep}

\begin{document}

\newcommand{\mycomment}[1]{}

\renewcommand{\figurename}{Fig.}

\title{Self-consistent context aware conformer transducer for speech recognition}

\author{Konstantin Kolokolov, Pavel Pekichev, Karthik Raghunathan \\
Webex Intelligence, Cisco Systems\\
771 Alder Dr, Milpitas, CA 95035\\
\{konstank, ppekiche, ktick\}@cisco.com
}

\maketitle
\thispagestyle{empty}

\begin{abstract}
 
We introduce a novel neural network module that adeptly handles recursive data flow in neural network architectures. At its core, this module employs a self-consistent approach where a set of recursive equations is solved iteratively, halting when the difference between two consecutive iterations falls below a defined threshold. Leveraging this mechanism, we construct a new neural network architecture, an extension of the conformer transducer, which enriches automatic speech recognition systems with a stream of contextual information. Our method notably improves the accuracy of recognizing rare words without adversely affecting the word error rate for common vocabulary. We investigate the improvement in accuracy for these uncommon words using our novel model, both independently and in conjunction with shallow fusion with a context language model. Our findings reveal that the combination of both approaches can improve the accuracy of detecting rare words by as much as 4.5 times. Our proposed self-consistent recursive methodology is versatile and adaptable, compatible with many recently developed encoders, and has the potential to drive model improvements in speech recognition and beyond.

\end{abstract}

\section{Introduction}
End-to-end speech-to-text systems have emerged as a dominant technology in voice recognition, surpassing the performance of the previously prevalent multi-component HMM-DNN systems. Among the various model architectures that stand out are: Connectionist Temporal Classification  \cite{Graves:06icml}, Listen-Attend-Spell \cite{LAS-2016}, Time-Depth Separable Convolutions \cite{hannun2019sequencetosequence}, Recurrent Neural Network Transducer \cite{graves-2012-sequen-trans}, Transformer \cite{Dong-no-recurrence-s2s-model-2018}, Conformer \cite{gulati2020conformer}, SqueezeFormer \cite{kim2022squeezeformer}, Fast Conformer\cite{rekesh2023fast}, and Zipformer \cite{yao2023zipformer}. These models all share an encoder-decoder architecture. 
One major limitation of E2E systems is their inability
to accurately recognize words that are either absent or infrequently present in the training data, for example, entity names, person names, etc. To address this issue,  two main strategies have been proposed: model-based and decoding-based approaches. The first approach modifies the model to integrate user-provided context words \cite{Pundak2018DeepCE,Chang2021ContextAwareTT, MysoreSathyendra2022,fu2023robust}, while the second approach applies an external context score using techniques like shallow fusion or  on-the-fly re-scoring \cite{Zhao2019end2end,wang2023contextual}.

\section{Model}

\subsection{Conformer transducer}

\begin{figure}
\centering
    \includegraphics[width=0.3\linewidth]{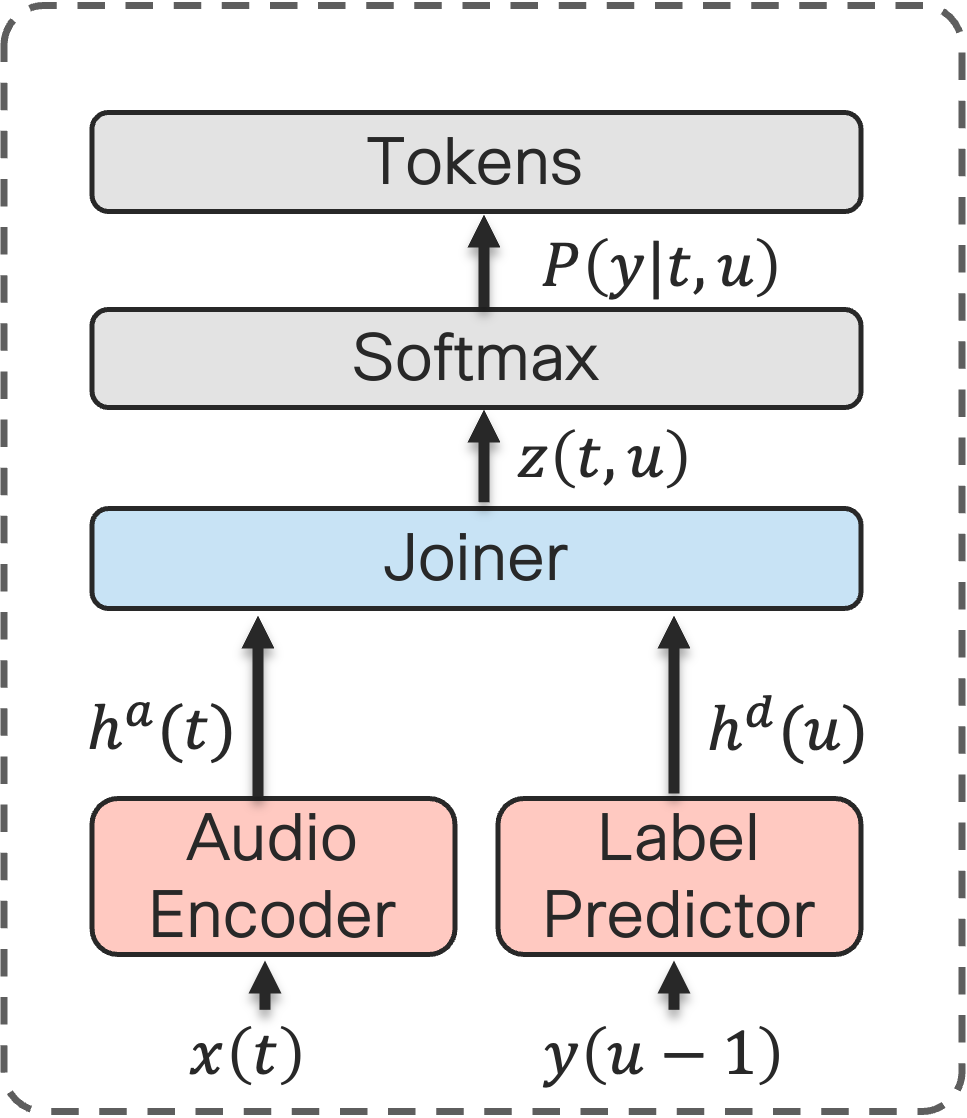}
  \caption{Transducer architecture.}
  \label{fig:transducer}
\end{figure}

Our model processes a set of audio features, denoted by $X$, and generates an utterance of tokens, represented by $Y$. The sequence-to-sequence model encodes the audio input $X$ into a  hidden vector representation using an encoder and then uses a decoder to predict the sequence of output tokens.
The transducer model consists of the three main modules: an audio encoder, a predictor (also referred to in the literature as a predictor network or decoder) and a joiner (or joint network), see \figurename{\ref{fig:transducer}}.

The audio encoder employs a series of stacked conformer layers \cite{gulati2020conformer}. 
For our predictor, we added a convolutional layer to a modified version of a predictor network \cite{Ghodsi2020StatelessPredictor}. The joint network then merges the audio encoder outputs $h^a$ with the predictor outputs $h^d$. The combined joiner output $z$ is fed into a softmax layer, resulting in a probability distribution over blank symbol and output tokens. We use sentence piece tokenization \cite{kudo2018sentencepiece} in our work. 

\subsection{Context conformer transducer}

\begin{figure}[H]
\centering
  \includegraphics[width=0.9\linewidth]{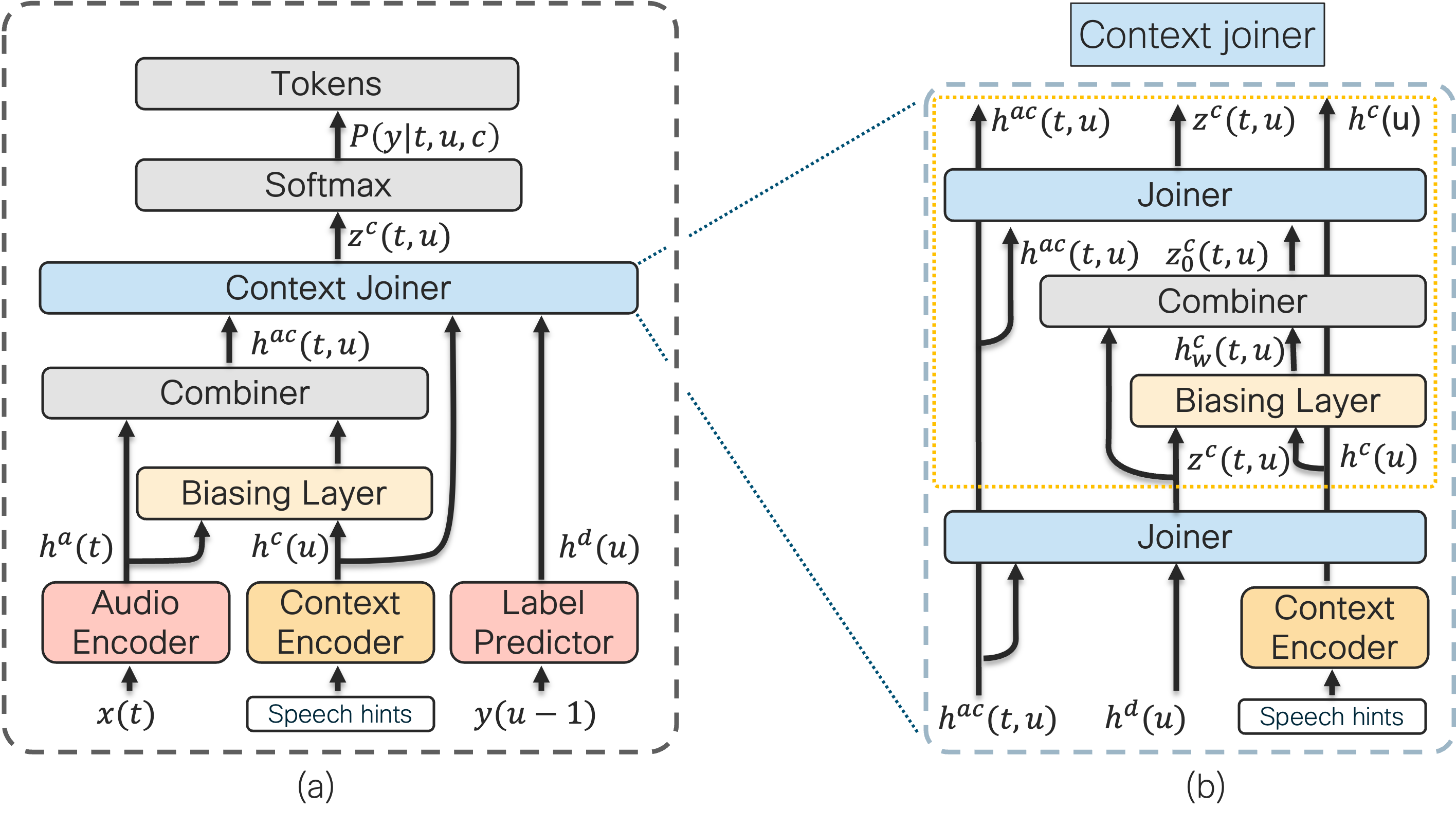}
  \caption{Context transducer architecture.}
  \label{fig:context_transducer}
\end{figure}

To enhance the Automatic Speech Recognition (ASR) model with contextual information, we add several components based on the Context-Aware Transducer Transformer architecture proposed in \cite{Chang2021ContextAwareTT}. These components include a context encoder, a biasing layer, and a combiner, as shown in \figurename{\ref{fig:context_transducer}}. The output of audio encoder is passing through biasing layer and combiner, the output of the combiner becomes the input of joiner.
Another pair of biasing layer and combiner is used in context joiner \figurename{\ref{fig:context_transducer}}(b). We will describe the data flow more in details later.

The context encoder is designed to handle named entities, such as user names, company names, or any custom terminology provided by the user - collectively termed as speech hints. It consists of an embedding layer and stacked BLSTM layers. First, the user-provided list of speech hints is tokenized into a sequence of tokens, which are then passed through the embedding layer and subsequently by the BLSTM layers. The last hidden state of the BLSTM serves as a vector representation for the embedded speech hints. The batch size of the BLSTM corresponds to the number of speech hints. The context encoder ultimately transforms the list of speech hints into a set of fixed-dimension embedding vectors. We use two sets of identical context encoders, combiners, and biasing layers, one for the audio encoder and another one for the joiner. The output of the corresponding context encoder is combined with the outputs of the audio encoder and joiner.

{\bf Audio Encoder:} First, let's consider the audio encoder, see \figurename{\ref{fig:context_transducer}}(a). The outputs of the audio encoder $h^a(t)$ and context encoder $h^c(u)$ are fed into the biasing layer. The latter is a multi-head attention layer where the audio encoder outputs serve as queries $Q$, and the context encoder outputs, which are the embedded vector representations of the speech hints, function as keys $K$ and values $V$. The attention is then computed using the following expression:

\begin{equation}
h^c_w = Softmax\left( \frac{QW_q(KW_k)^T}{\sqrt{d}} \right)VW_v
\label{eq:cross_attention}
\end{equation}

Here, $W_q$, $W_k$, and $W_v$ are learnable parameters, $Q$ is $h^a(t)$, $K$ and $V$ are $h^c(u)$, and $d$ is the dimension of the keys. The resulting scores from the softmax layer are used to determine weighted contribution from the speech hints embeddings. This value is then fed to the combiner layer that consists of two normalization layers, concatenation and a subsequent linear projection layer:

\begin{equation}
\renewcommand*{\arraystretch}{1.4}
\begin{array}{@{}l@{}}
h^{cat} = Concat \left(LayerNorm(h^a), LayerNorm(h^c_w) \right)\\
h^{ac} = Linear(h^{cat})
\end{array}
\end{equation}

{\bf Label Predictor:} Integrating context encoder modules into the label predictor was considered but ultimately not implemented for the following reasons. The label predictor essentially functions like an internal language model: given a sequence of previously generated tokens $y[u-m, \ldots, u-1]$, it generates a hidden vector which, with the addition of a softmax layer, would resemble a language model architecture. If such a neural network has enough learnable parameters and the appropriate architecture, it can predict the next token or word. However, our label predictor is quite simple, consisting only of an embedding vector and a convolution layer, making accurate next token prediction unlikely. The scores from the softmax layer in Eq.\eqref{eq:cross_attention} determines weighted contributions from the speech hints. In other words we get the probability vector to attend a particular speech hint. If we add context encoder modules to our label predictor, we will get the following task that is similar to language model: given a sequence of previously generated tokens and the label predictor layer we want to predict weighted contributions from the speech hints. If the predictor is poor in quality the result of contributions speech hints  is poor in quality as well. This is supported by our experiments.

{\bf Context Joiner:} We found that pairing the context encoder with the label predictor is ineffective due to the inability to properly attend to the speech hints embeddings. However, if we use the output of the joiner, which is the hidden representation of all previously generated tokens at the current timestamp $t$, we have a better chance of accurately attending to speech hints. This allows us to determine which speech hint is most relevant at the moment. The resulting attention from the speech hints can then be combined with the joiner's output, and fed back as an input to the joiner, instead of the output from the label predictor, as depicted in \figurename{\ref{fig:context_transducer}}(b).

This is expressed in the following set of equations:

\begin{equation}
\left\{
\renewcommand*{\arraystretch}{1.5}
\begin{array}{@{}l@{}}
h^c_w(t, u) = BiasingLayer\left(z^c(t,u), h^c(u) \right) \\
z^c_0(t,u) = Combiner\left(z^c(t,u), h^c_w(t, u)\right) \\
z^c(t,u) = Joiner\left(h^{ac}(t, u), z^c_0(t,u) \right) 
\end{array} 
\right. \
\label{eq:context_joiner_eq}
\end{equation}

Here, $h^{ac}(t, u)$ is the audio encoder and audio context combiner output, $z^c_0(t,u)$ is the joiner combiner output, $z^c(t,u)$ is the joiner output, and $h^c(u)$ is the context encoder output. Initially, $z^c(t,u)$ is equal to the predictor output $h^{d}(u)$ (see lower part of \figurename{\ref{fig:context_transducer}}(b)). The biasing layer in context joiner \eqref{eq:context_joiner_eq} is also described by equation \eqref{eq:cross_attention}. $W_q$, $W_k$, and $W_v$ are another set of learnable parameters, $Q$ is $z^c(t,u)$, $K$ and $V$ are $h^c(u)$.

We get a set of self-recurrent equations in \eqref{eq:context_joiner_eq}, as the first input argument of the biasing layer is the output of the joiner, and the second input argument of the joiner is the output of the combiner, which in turn depends on the biasing layer output.

To solve the above mentioned problem, we draw inspiration from a self-consistent approach commonly used in semiconductor physics. This approach involves iteratively solving a set of coupled equations, such as the Schrödinger equation in tandem with the Poisson equation:

\begin{equation}
  \left\{
  \renewcommand*{\arraystretch}{2.4}
  \begin{array}{@{}l@{}}
    -\dfrac{\partial^2 F(x)}{\partial x^2} + V(x)F(x)=EF(x)\\
    \phantom{-}\dfrac{\partial^2 V(x)}{\partial x^2}=\left|F(x)\right|^2
  \end{array}\right.\,
\end{equation}

Here, $E$ is the energy, $F$ is the wave-function and $V$ is the band profile. For simplicity, we've omitted physical constants like electron charge. The goal is to find $F$ and $V$.  We can't solve these equations directly in tandem, but we can solve them iteratively. We start by assuming  $V(x) \equiv 0$ to solve the first equation for $F(x)$, then use $F(x)$ to find $V(x)$, and iterate the process. We repeat this procedure until both $V(x)$ and $F(x)$ converge to their asymptotes, i.e. when the differences between iterations fall below a predefined threshold. This iterative method is known as self-consistent calculations, which we will adapt to our task.

We can apply this technique to solve the equations \eqref{eq:context_joiner_eq}. First, we take $h^{ac}(t, u)$ and $h^{d}(u)$ as inputs and compute the joiner output $z^c(t,u)$. Then we compute the biasing layer output $h^c_w(t,u)$ and combiner output $z^c_0(t,u)$ and feed it to the joiner again. We repeat this process until the difference between the joiner outputs from consecutive iterations $n$ and $n-1$ is some small value. In practice, convergence is typically achieved within three iterations, as shown in Table \ref{table: iters_comparison}. The table displays the averaged maximum difference and mean difference between joiner outputs across consecutive iterations, as calculated with a 95\% confidence interval over five independent trials.

\begin{table}
  \caption{Convergence of self-consistent iterations}
  \centering
  \begin{tabular}{cll}
    \toprule
    {\bf Iteration number}&{\bf Avg max diff} & {\bf Avg mean diff}\\
    \midrule
    1	& $(3.85  \pm 0.11) \times 10^{+6}$ & ($\phantom{-} 1.35  \pm  0.04)  \times 10^{-2}$	\\
    2	& $(2.43  \pm 0.07)	\times 10^{+1}$ & ($-3.99 \pm  0.11)  \times 10^{-8}$	\\
    3	& $(2.11  \pm 0.06) \times 10^{-1}$ & ($-8.99 \pm 0.25) \times 10^{-12}$	\\
    4	& $(2.19  \pm 0.06) \times 10^{-2}$ & ($-7.55 \pm 0.21) \times 10^{-13}$	\\
    5	& $(1.15  \pm 0.03) \times 10^{-2}$ & ($\phantom{-}1.29 \pm 0.04) \times 10^{-12}$	\\
    6	& $(1.03  \pm 0.03) \times 10^{-2}$ & ($-3.12 \pm 0.09) \times 10^{-13}$	\\
    \bottomrule
  \end{tabular}
  \label{table: iters_comparison}
\end{table}

The detailed rolled out diagram of the data flow when $n$ is equal to 1 is shown in \figurename{\ref{fig:context_transducer}}(b). For $n > 1$ we need to repeat the steps within the orange dotted block. The pseudo-code for this algorithm is outlined in Algorithm \ref{algorithm: self-cons-joiner}.

\begin{algorithm}
	\caption{Context joiner as self-consistent recursive module}
	\begin{algorithmic}[1]
        \State {Let $h^{ac}(t, u)$ be acoustic encoder output (or acoustic context combiner output)}
        \State {Let $h^{d}(u)$ be predictor (decoder) output}
        \State {Let $h^c(u)$ be context encoder embeddings}
        \State {Let $N$ be maximum iteration number}
        \State {Let $th$ be a threshold for consecutive iterations stop}
        \State {$z^c(t,u)$ = Joiner$\left(h^{ac}(t), h^{d}(u) \right)$}
        \State {$z^c_{prev}(t,u)$ = $z^c(t,u)$}
        \State {$\Delta z^c$ = $ +\infty$}
		\For {$n$ = 1 to $N$}
        \State {$h^c_w(t, u)$ = BiasingLayer$\left(z^c(t,u), h^c(u)\right)$}
        \State {$z^c_0(t,u)$ = Combiner$\left(z^c(t,u), h^c_w(t, u)\right)$}
        \State {$z^c(t,u)$ = Joiner$\left(h^{ac}(t), z^c_0(t,u) \right)$}
        \State {$\Delta z^c$ = $abs(mean(z^c(t,u) - z^c_{prev}(t,u)))$}
        \If {$n > 1$ and $\Delta z^c$ < $th$}
            \State {break loop}
    	\EndIf
        \State {$z^c_{prev}(t,u)$ = $z^c(t,u)$}
        \EndFor
        \State {return $z^c(t,u)$}
	\end{algorithmic} 
    \label{algorithm: self-cons-joiner}
\end{algorithm} 

\section{Experimental setup}

Our experimental setup utilized approximately 12,000 hours of speech audio from an in-house anonymized dataset for training the model, with an additional 40 hours set aside for evaluation. To train our model, we generated synthetic speech hints on-the-fly using the following procedure:
\begin{itemize}
\setlength{\itemindent}{0em}
\setlength\itemsep{0.5em}
  \item We randomly select a set of words from the provided ground truth text. Each word is then augmented by duplicating one or more characters. For example, the word {\it cat} could be transformed to {\it catt},  {\it triple} might become {\it tripple}, etc.
  \item We randomly augment the selected speech hints words to produce similar-sounding variants. For example, we swap {\it k} $\Leftrightarrow$ {\it c}, {\it j} $\Leftrightarrow$ {\it g}, etc. The extent of change varies with some words changing only one letter and others experiencing multiple changes. E.g., {\it cat} $\Rightarrow$ {\it kat}, {\it ginger} $\Rightarrow$ {\it jinger}, {\it cereal} $\Rightarrow$ {\it serial}, {\it hay} $\Rightarrow$ {\it hey}, etc.
\end{itemize}

We create three types of samples:
  \begin{itemize}\setlength{\itemindent}{0em}
  \setlength\itemsep{0.5em}
    \item Original samples, representing scenarios where no speech hints are provided.
    \item Samples containing only negative speech hints, to account for cases where the speech hints do not appear in the ground truth.
    \item Samples with a mix of positive and negative speech hints, reflecting situations where only some of the speech hints are present in the ground truth.
  \end{itemize}

The architecture of our model's encoder features 12 causal conformer layers with a hidden dimension of 256 and a feedforward dimension of 2048. The predictor consists of an embedding layer and a causal convolution layer with a kernel size of 3. For the context encoder, we employ a two-layer BLSTM with a dimension of 256. The multi-head cross-attention layer within the context biasing layer has 8 heads. A shared embedding layer was used across the model in both the predictor and context encoders. The number of tokens in the model was 500.

Our model had around 85.2M trainable parameters. We train the model with a pruned transducer loss function \cite{kuang2022pruned}. The model training was conducted using the K2 icefall \cite{k2_icefall} and PyTorch Distributed Data Parallel (DDP) packages. It took about 8 days to train the model on a setup consisting of two nodes with 8 Nvidia V100 GPUs with 32GB of memory. Each node was run with 8 workers and used the adaptive batch size feature implemented in K2 icefall.

\section{Results and discussion}
We evaluated the results of our experiments using relative word error rate reduction (WERR) and out-of-vocabulary word recognition accuracy (OOV accuracy). WERR is calculated by comparing the change in word error rate (WER) when speech hints are included versus when they are absent. OOV accuracy is determined by the proportion of correctly transcribed speech hints in an utterance relative to the total number of speech hints present.

To create speech hints for the test dataset, we first identified words from the test set that were not present in the training data. We then selected words that only appear once in the test set. The number of speech hints evaluated ranged from 100 to 1,000.

\begin{table}
  \caption{WERR and OOV accuracy with different ways of providing speech hints}
  \centering
  \begin{tabularx}{\columnwidth}{ X p{1.8cm} p{1.8cm} p{1.8cm}  }
    \toprule
    {\bf Model configuration}&{\bf Number of speech hints} &{\bf WERR, \%} & {\bf OOV accuracy, \%}\\
    \midrule
    No hints                           & 0   & \phantom{-}0     & 8.3   \\
    \midrule
    Context encoder                    & 100 & \phantom{-}0.76  & 12.03 \\
    Shallow fusion                     & 100 & -1.33 & 33.08 \\
    Context encoder + Shallow fusion   & 100 & -2.32 & 45.11 \\
    \midrule
    Context encoder                    & 200 & \phantom{-}0.95  & 10.68 \\
    Shallow fusion                     & 200 & -2.39 & 36.89 \\
    Context encoder + Shallow fusion   & 200 & -0.08 & 41.42 \\
    \midrule
    Context encoder                    & 500 & \phantom{-}3.50  & 14.19 \\
    Shallow fusion                     & 500 & -3.99 & 35.33 \\
    Context encoder + Shallow fusion   & 500 & \phantom{-}0.46  & 36.8  \\
    \midrule
    Context encoder                    & 1000 & \phantom{-}3.00  & 14.52\\
    Shallow fusion                     & 1000 & -4.37 & 35.01\\
    Context encoder + Shallow fusion   & 1000 & -1.41 & 35.21\\
    \bottomrule
  \end{tabularx}
  \label{table: werr_and_oov_acc}
\end{table}

The results of experiments are shown in Table \ref{table: werr_and_oov_acc}. The first row shows the results of the baseline experiment when no speech hints are provided (WERR is zero). The subsequent rows detail the outcomes when speech hints are provided solely to the context encoder, used solely in shallow fusion, and when both techniques are employed. The results are reported for varying numbers of speech hints: 100, 200, 500, and 1,000. For the scenario with 100 speech hints, we observed that incorporating speech hints within the context encoder led to about a 50\% relative increase in OOV accuracy over the baseline. Shallow fusion showed even more substantial improvements, raising our OOV accuracy from 8.3\% to 33\%. Combining both context encoder and shallow fusion techniques results in the best OOV accuracy at 45\%. The relative WER change does not exceed 2.3\% in across any of these experiments. This pattern holds true as the number of speech hints increases, demonstrating that using both the context encoder architecture and shallow fusion significantly enhances OOV accuracy while maintaining a relatively stable WER.

The proposed idea of a self-consistent recursive neural network module that we developed for context joiner can be extended to a variety of scenarios involving recursive data flow within neural networks. It is not limited to the speech domain; it can be incorporated into any application relying on neural network architectures.

For instance, the self-consistent recursive module can be used to integrate a custom dictionary into a language model. In \figurename{\ref{fig:context_transducer}}(b), we can remove the joiner as well all the $h^{ac}(t)$ arrows that are related to the acoustic part of the model. We are then left with a self-consistent recursive module that is comprised of a biasing layer and a combiner, where the combiner's input is the output of the biasing layer and vice versa:

\begin{equation}
\left\{
\renewcommand*{\arraystretch}{1.5}
\begin{array}{@{}l@{}}
h^c_w(u) = BiasingLayer\left(z^c(u), h^c(u) \right) \\
z^c(u) = Combiner\left(z^c(u), h^c_w(u)\right) 
\end{array} 
\right. \
\label{eq:self_recursive_eq}
\end{equation}

Here, $z^c(u)$ is the combiner output, $h^c(u)$ is the context encoder output, and $h^c_w(u)$ is the output of biasing layer or cross-attention (see Eq.\eqref{eq:cross_attention}). The algorithm for the language model case could be derived from Algorithm \ref{algorithm: self-cons-joiner}, where $h^{ca}(t,u)$ can be removed and the joiner layer can be replaced with an identity layer. This recursive module for custom dictionary can be added to an already pretrained language model. In that case, all network parameters, except the self-consistent recursive module, are initialized with a pretrained model \cite{MysoreSathyendra2022}, which  avoids the need to train the language model with custom dictionary from scratch.

We also found in our experiments that adding layer normalization to the context encoder output leads to a speed up of training process. Layer normalization allows the attention mechanism to create an attention query that attends to all keys equally and prevents keys, or in our case speech hints, from being overlooked \cite{brody2023expressivity}. Another set of experiments showed that the context joiner module is compatible with the Zipformer speech-to-text architecture \cite{yao2023zipformer} and using data from publicly available LibriSpeech dataset \cite{7178964-LibriSpeech} for training provides similar results in enhancement in the recognition of uncommon words.

\section{Conclusion}

We introduced a novel neural network module specifically designed to handle recursive data flow within neural networks. This module was successfully incorporated into a new architecture of the conformer-transducer based automatic speech recognition (ASR) model, which includes contextual data. Our results demonstrate that when this architectural approach is combined with shallow fusion using a context language model, there is a notable enhancement in the recognition of uncommon words.

Furthermore, the versatility of the proposed architecture is one of its key strengths, as it is not constrained to any single encoder design. It can be readily adapted to work with a range of recently developed encoders, including SqueezeFormer \cite{kim2022squeezeformer}, Fast Conformer\cite{rekesh2023fast} and Zipformer \cite{yao2023zipformer}. The potential for broader application suggests that our approach could be influential in improving models in speech recognition and beyond.

\bibliography{context_encoder_paper}

\begin{thebibliography}{10}

\bibitem{Graves:06icml}
A.~Graves, S.~Fernandez, F.~Gomez, and J.~Schmidhuber.
\newblock Connectionist temporal classification: Labelling unsegmented sequence data with recurrent neural nets.
\newblock In {\em ICML '06: Proceedings of the International Conference on Machine Learning}, 2006.

\bibitem{LAS-2016}
William Chan, Navdeep Jaitly, Quoc~V. Le, and Oriol Vinyals.
\newblock Listen, attend and spell: A neural network for large vocabulary conversational speech recognition.
\newblock In {\em ICASSP}, 2016.

\bibitem{hannun2019sequencetosequence}
Awni Hannun, Ann Lee, Qiantong Xu, and Ronan Collobert.
\newblock Sequence-to-sequence speech recognition with time-depth separable convolutions.
\newblock {\em CoRR}, abs/1904.02619, 2019.

\bibitem{graves-2012-sequen-trans}
A.~{Graves}.
\newblock {Sequence Transduction With Recurrent Neural Networks}.
\newblock {\em ArXiv e-prints}, November 2012.

\bibitem{Dong-no-recurrence-s2s-model-2018}
Linhao Dong, Shuang Xu, and Bo~Xu.
\newblock Speech-transformer: A no-recurrence sequence-to-sequence model for speech recognition.
\newblock In {\em 2018 IEEE International Conference on Acoustics, Speech and Signal Processing (ICASSP)}, pages 5884--5888, 2018.

\bibitem{gulati2020conformer}
Anmol Gulati, James Qin, Chung-Cheng Chiu, Niki Parmar, Yu~Zhang, Jiahui Yu, Wei Han, Shibo Wang, Zhengdong Zhang, Yonghui Wu, and Ruoming Pang.
\newblock Conformer: Convolution-augmented transformer for speech recognition.
\newblock {\em ArXiv e-prints}, 2020.

\bibitem{kim2022squeezeformer}
Sehoon Kim, Amir Gholami, Albert Shaw, Nicholas Lee, Karttikeya Mangalam, Jitendra Malik, Michael~W. Mahoney, and Kurt Keutzer.
\newblock Squeezeformer: An efficient transformer for automatic speech recognition.
\newblock {\em ArXiv e-prints}, 2022.

\bibitem{rekesh2023fast}
Dima Rekesh, Nithin~Rao Koluguri, Samuel Kriman, Somshubra Majumdar, Vahid Noroozi, He~Huang, Oleksii Hrinchuk, Krishna Puvvada, Ankur Kumar, Jagadeesh Balam, and Boris Ginsburg.
\newblock Fast conformer with linearly scalable attention for efficient speech recognition.
\newblock {\em ArXiv e-prints}, 2023.

\bibitem{yao2023zipformer}
Zengwei Yao, Liyong Guo, Xiaoyu Yang, Wei Kang, Fangjun Kuang, Yifan Yang, Zengrui Jin, Long Lin, and Daniel Povey.
\newblock Zipformer: A faster and better encoder for automatic speech recognition.
\newblock {\em ArXiv e-prints}, 2023.

\bibitem{Pundak2018DeepCE}
Golan Pundak, Tara~N. Sainath, Rohit Prabhavalkar, Anjuli Kannan, and Ding Zhao.
\newblock Deep context: End-to-end contextual speech recognition.
\newblock {\em 2018 IEEE Spoken Language Technology Workshop (SLT)}, pages 418--425, 2018.

\bibitem{Chang2021ContextAwareTT}
Feng-Ju Chang, Jing Liu, Martin~H. Radfar, Athanasios Mouchtaris, Maurizio Omologo, Ariya Rastrow, and Siegfried Kunzmann.
\newblock Context-aware transformer transducer for speech recognition.
\newblock {\em 2021 IEEE Automatic Speech Recognition and Understanding Workshop (ASRU)}, pages 503--510, 2021.

\bibitem{MysoreSathyendra2022}
Kanthashree~Mysore Sathyendra, Thejaswi Muniyappa, Feng-Ju~(Claire) Chang, Jing Liu, Jinru Su, Grant Strimel, Thanasis Mouchtaris, and Siegfried Kunzmann.
\newblock Contextual adapters for personalized speech recognition in neural transducers.
\newblock In {\em ICASSP 2022}, 2022.

\bibitem{fu2023robust}
Xuandi Fu, Kanthashree~Mysore Sathyendra, Ankur Gandhe, Jing Liu, Grant~P. Strimel, Ross McGowan, and Athanasios Mouchtaris.
\newblock Robust acoustic and semantic contextual biasing in neural transducers for speech recognition.
\newblock {\em ArXiv e-prints}, 2023.

\bibitem{Zhao2019end2end}
Ding Zhao, Tara~N. Sainath, David Rybach, Pat Rondon, Deepti Bhatia, Bo~Li, and Ruoming Pang.
\newblock {Shallow-Fusion End-to-End Contextual Biasing}.
\newblock In {\em Proc. Interspeech 2019}, pages 1418--1422, 2019.

\bibitem{wang2023contextual}
Weiran Wang, Zelin Wu, Diamantino Caseiro, Tsendsuren Munkhdalai, Khe~Chai Sim, Pat Rondon, Golan Pundak, Gan Song, Rohit Prabhavalkar, Zhong Meng, Ding Zhao, Tara Sainath, and Pedro~Moreno Mengibar.
\newblock Contextual biasing with the knuth-morris-pratt matching algorithm.
\newblock {\em ArXiv e-prints}, 2023.

\bibitem{Ghodsi2020StatelessPredictor}
Mohammadreza Ghodsi, Xiaofeng Liu, James Apfel, Rodrigo Cabrera, and Eugene Weinstein.
\newblock Rnn-transducer with stateless prediction network.
\newblock In {\em ICASSP 2020 - 2020 IEEE International Conference on Acoustics, Speech and Signal Processing (ICASSP)}, pages 7049--7053, 2020.

\bibitem{kudo2018sentencepiece}
Taku Kudo and John Richardson.
\newblock Sentencepiece: A simple and language independent subword tokenizer and detokenizer for neural text processing.
\newblock {\em ArXiv e-prints}, 2018.

\bibitem{kuang2022pruned}
Fangjun Kuang, Liyong Guo, Wei Kang, Long Lin, Mingshuang Luo, Zengwei Yao, and Daniel Povey.
\newblock Pruned rnn-t for fast, memory-eﬀicient asr training.
\newblock In {\em Proc. Interspeech 2022}, pages 2068--2072, 2022.

\bibitem{k2_icefall}
K2 icefall.
\newblock \url{https://github.com/k2-fsa/icefall}, 2023.

\bibitem{brody2023expressivity}
Shaked Brody, Uri Alon, and Eran Yahav.
\newblock On the expressivity role of layernorm in transformers' attention.
\newblock {\em ArXiv e-prints}, 2023.

\bibitem{7178964-LibriSpeech}
Vassil Panayotov, Guoguo Chen, Daniel Povey, and Sanjeev Khudanpur.
\newblock Librispeech: An asr corpus based on public domain audio books.
\newblock In {\em 2015 IEEE International Conference on Acoustics, Speech and Signal Processing (ICASSP)}, pages 5206--5210, 2015.

\end{thebibliography}

\end{document}